\documentclass{article}
\usepackage{ijcai26}

\usepackage{times}
\usepackage{soul}
\usepackage{url}
\usepackage[hidelinks]{hyperref}
\usepackage[utf8]{inputenc}
\usepackage[small]{caption}
\usepackage{graphicx}
\usepackage{amsmath}
\usepackage{amsthm}
\usepackage{booktabs}
\usepackage{algorithm}
\usepackage{algorithmic}
\usepackage[switch]{lineno}
\usepackage{multirow}
\usepackage{amssymb}
\usepackage[most]{tcolorbox}
\usepackage{xcolor}
\usepackage{pifont}
\usepackage{minted}
\usepackage{tabularx}

\title{HyEvo: Self-Evolving Hybrid Agentic Workflows for Efficient Reasoning}

\author{
Beibei Xu$^1$\and
Yutong Ye$^2$\and
Chuyun Shen$^3$\and
Yingbo Zhou$^4$\and
Cheng Chen$^1$\and
Mingsong Chen$^1$\\
\affiliations
$^1$East China Normal University\and
$^2$Beihang University\\
$^3$Shanghai University of International Business and Economics\and
$^4$Fudan University\\
\emails
xubeibei0814@126.com
}

\begin{document}

\maketitle

\begin{abstract}
Although agentic workflows have demonstrated strong potential for solving complex tasks, existing automated generation methods remain inefficient and underperform, as they rely on predefined operator libraries and homogeneous LLM-only workflows in which all task-level computation is performed through probabilistic inference. To address these limitations, we propose HyEvo, an automated workflow-generation framework that leverages heterogeneous atomic synthesis. HyEvo integrates probabilistic LLM nodes for semantic reasoning with deterministic code nodes for rule-based execution, offloading predictable operations from LLM inference and reducing inference cost and execution latency. To efficiently navigate the hybrid search space, HyEvo employs an LLM-driven multi-island evolutionary strategy with a reflect-then-generate mechanism, iteratively refining both workflow topology and node logic via execution feedback. Comprehensive experiments show that HyEvo consistently outperforms existing methods across diverse reasoning and coding benchmarks, while reducing inference cost and execution latency by up to 19$\times$ and 16$\times$, respectively, compared to the state-of-the-art open-source baseline.

\end{abstract}

\section{Introduction}

Agentic workflows have emerged as a promising paradigm for solving complex tasks by enabling Large Language Models (LLMs) to overcome the limitations of single-pass inference through structured decomposition~\cite{ijcai2024p890}. By organizing specialized nodes into compositional graph topologies, these systems enable reasoning capabilities that direct prompting often fails to achieve. However, constructing such workflows remains challenging, as identifying appropriate granularity and topology relies on labor-intensive manual design that is difficult to scale~\cite{ijcai2024p3}.

\begin{figure}[!t]
    \centering
    \includegraphics[width=0.9\linewidth]{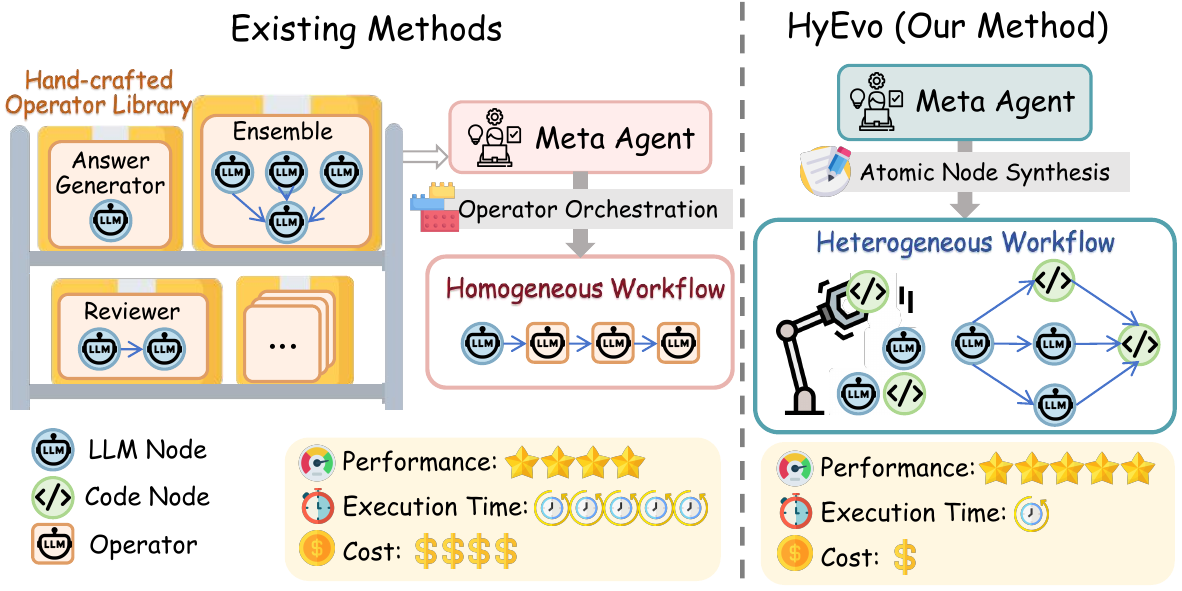}
    \vspace{-0.05in}
    \caption{Comparison between existing methods and HyEvo.}
    \vspace{-0.1in}
    \label{fig:vs}
\end{figure}

To address this bottleneck, automated workflow generation methods have been proposed, typically employing a meta-agent to replace human experts in planning and making decisions~\cite{li2025review,google_maas,huautomated}. However, as shown in the left part of Figure~\ref{fig:vs}, most existing approaches adopt an orchestration paradigm, where the meta-agent selects from a predefined library of operators or configures LLM parameters~\cite{zhang2025aflow,zhang2025evoflow,ijcai2025p296,su2025debflow,zhangmulti}. This approach is not fully automated because the operator library itself still requires manual crafting. Furthermore, it results in workflows composed entirely of LLM nodes. We define these structures as homogeneous because all task-level semantic computation is performed through probabilistic LLM inference, without deterministic execution counterparts. This structural homogeneity leads to prohibitive execution overheads in both latency and cost.

To address this inefficiency, workflows should delegate deterministic sub-tasks to executable components instead of relying solely on probabilistic inference. Motivated by this observation, we propose HyEvo. As shown in the right part of Figure~\ref{fig:vs}, HyEvo can autonomously construct workflows by synthesizing two distinct types of atomic units: \textit{i) LLM nodes} responsible for complex semantic reasoning, and \textit{ii) code nodes} executing deterministic logic. Specifically, HyEvo delegates rule-based sub-tasks to code nodes that execute deterministic logic under explicit constraints, while LLM nodes are reserved for semantic reasoning. This heterogeneous design combines the deterministic efficiency of code nodes alongside the flexible reasoning capabilities of LLM nodes, ultimately improving task performance and significantly reducing inference costs and execution latency.

This autonomous generation capability inherently expands the search space, making it challenging for standard methods to identify optimal solutions. To efficiently navigate this hybrid landscape, we employ a multi-island evolutionary strategy. By maintaining diverse populations, this approach ensures a broad exploration of potential designs. To improve the efficiency of the evolutionary search process, we introduce a cascaded sandbox evaluation protocol that rapidly filters out suboptimal candidates with minimal computational cost. Furthermore, through a reflect-then-generate mechanism, the meta-agent analyzes execution feedback to diagnose shortcomings in both architectural design and node implementation. This integrates reasoning into the evolutionary process, enabling an intelligent and directed search for high-performing hybrid agentic workflows.

This paper mainly makes the following three contributions:
\begin{enumerate}
    \item 
    We propose HyEvo, a framework that automatically generates hybrid workflows for efficient reasoning. It integrates LLM and code as task-level nodes, combining semantic reasoning with deterministic execution.
    \item 
    We design an LLM-driven multi-island evolutionary strategy for HyEvo that employs a reflect-then-generate mechanism to efficiently navigate the expansive search space and discover high-performing workflows.
    \item
    We conduct comprehensive experiments on five well-known mathematical and coding benchmarks, demonstrating that HyEvo significantly outperforms state-of-the-art baselines in both performance and efficiency.
\end{enumerate}

\section{Related Work}
\paragraph{Agentic Systems and Workflows.}
To address the limitations of LLMs in tackling complex problems, research has shifted from single-turn reasoning to structured, multi-agent systems. Early prompting strategies, such as Chain-of-Thought (CoT)~\cite{wei2022chain} and Self-Consistency (SC)~\cite{wang2023selfconsistency}, enhanced reasoning depth but remained constrained by their implicit and linear reasoning paths. Consequently, researchers proposed agentic systems where agents collaborate through explicit coordination frameworks~\cite{NeurIPS2023camel,wu2024autogen,hong2024metagpt,ye2025masgpt}. Building on these frameworks, agentic workflows have emerged to model the reasoning process as an executable computation graph. Within this paradigm, researchers have manually crafted specialized topological structures to handle distinct tasks, ranging from role-playing and dialectical debates to embodied control and reflective loops~\cite{multi-persona,arXiv2023_MultiAgent-Debate,chen2024agentverse,ijcai2024p711,ijcai2025p14}. However, these hand-crafted designs are labor-intensive and generalize poorly across diverse domains.

\paragraph{Automatic Optimization of Agentic Workflows.}
To overcome the complexity of manual design, recent methods have been proposed to automate the generation of agentic workflows. ADAS~\cite{huautomated} formulates workflow generation as a search process within an unconstrained code space. However, navigating such a vast and unstructured space via linear mutation leads to inefficient exploration and unstable convergence. To mitigate these efficiency issues, subsequent studies~\cite{zhang2025aflow,gao2025flowreasonerreinforcingquerylevelmetaagents,wang2025scoreflow,su2025debflow} restrict the search space through manually predefined operators. Although these designed modules improve execution reliability, they inherently limit the degree of automation and constrain the solution space to compositions of high-level blocks. This coarse granularity hinders the refinement of fine-grained interaction logic and limits the discovery of novel atomic structures. Moreover, existing methods predominantly orchestrate LLM-based components, largely overlooking the systematic integration of deterministic code nodes for efficient execution.

\paragraph{Evolutionary Discovery of Programs.}
The study of evolutionary program discovery provides a broader foundation for automated workflow generation. Recent works~\cite{FunSearch2023,novikov2025alphaevolve,openevolve} demonstrate that evolutionary algorithms can evolve executable code to discover novel human-competitive algorithms, highlighting the creative potential of evolutionary search. However, these methods target isolated functions rather than structured agentic workflows. Consequently, some recent studies have attempted to apply evolutionary algorithms to the generation of agentic workflows~\cite{zhang2025evoflow,zheng2025mermaidflow}. Nevertheless, to simplify the modeling, these methods typically restrict the search space to predefined operators. This dependence not only limits flexibility but also fails to fully eliminate the need for manual design. Furthermore, such approaches predominantly generate homogeneous workflows composed solely of LLM nodes, resulting in severe inefficiencies and limited performance.

To the best of our knowledge, HyEvo is the first framework to autonomously construct heterogeneous workflows comprising both LLM and code nodes, achieved by extending evolutionary search from operator orchestration to atomic node synthesis. Based on our multi-island MAP-Elites strategy and reflect-then-generate mechanism, HyEvo not only outperforms state-of-the-art baselines but also significantly reduces inference costs and execution latency.

\section{Preliminaries}

\begin{figure*}[t]
    \centering
    \includegraphics[width=0.85\textwidth]{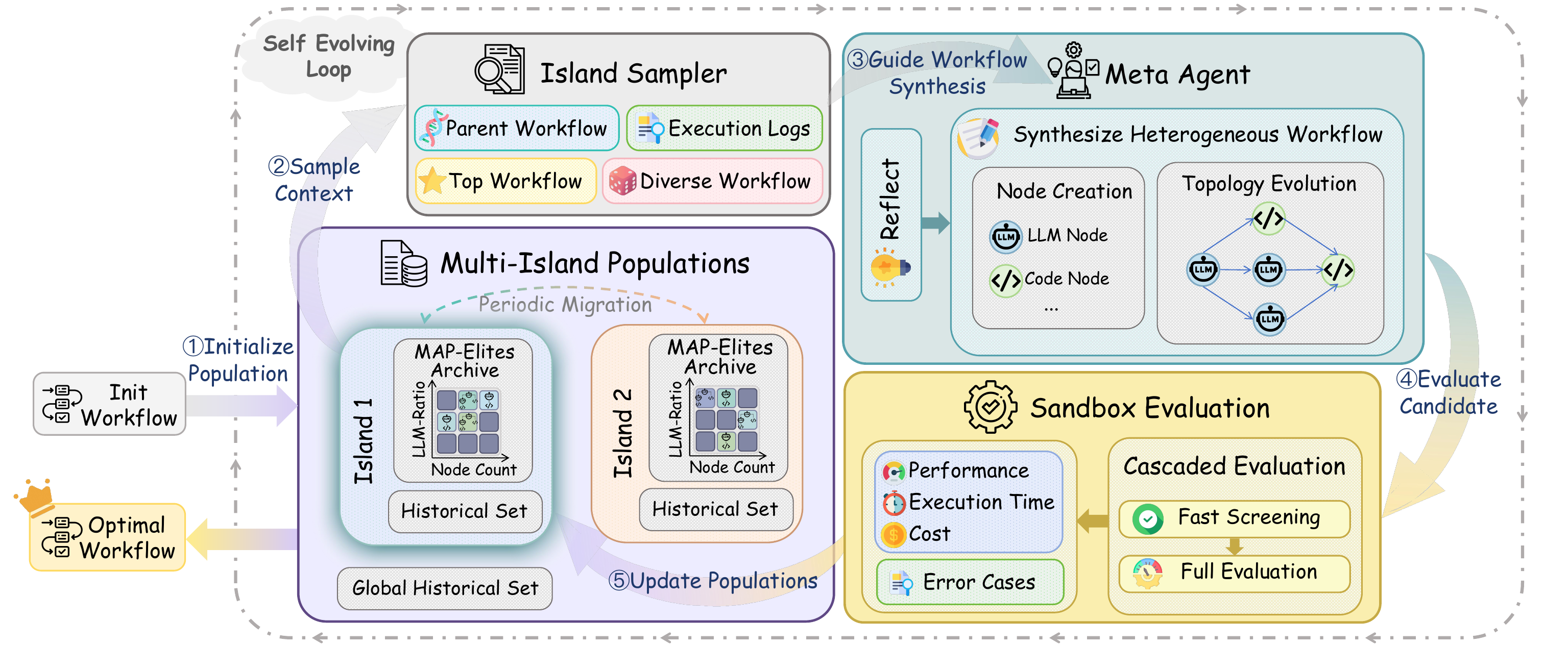}
        \vspace{-0.1in}
    \caption{Overview of the HyEvo framework}
    \vspace{-0.1in}
    \label{fig:method}
\end{figure*}


\paragraph{The Hybrid Search Space.}
We distinguish nodes by their semantic execution roles and formally define the hybrid node space $\mathcal{V}$ as the union of two generative subspaces: $\mathcal{V} = \mathcal{V}_{\text{LLM}} \cup \mathcal{V}_{\text{Code}}$. The LLM node $v^{\text{LLM}} \in \mathcal{V}_{\text{LLM}}$ functions as a probabilistic reasoning unit:
\begin{equation}
    v^{\text{LLM}} = \langle m, \mathcal{P}, \tau \rangle,
\end{equation}
where $m$ is the backbone model, $\mathcal{P}$ denotes the instruction generated by the meta-agent within the text space, and $\tau \in [0, 1]$ controls the sampling temperature. Meanwhile, distinct from existing methods that rely solely on LLM nodes, we introduce task-level code nodes $v^{\text{Code}} \in \mathcal{V}_{\text{Code}}$ to support deterministic execution for rule-based sub-tasks:
\begin{equation}
    v^{\text{Code}} = \langle c_{\text{src}}, \mathcal{I}_{\text{sig}} \rangle, \quad \text{s.t. } \Phi(c_{\text{src}}, x) \to y_{\text{det}},
\end{equation}
where $c_{\text{src}}$ is the synthesized source code, $\mathcal{I}_{\text{sig}}$ is the I/O type signature ensuring compatibility within the workflow graph, and $\Phi$ denotes the deterministic execution function mapping input $x$ to a strictly reproducible output $y_{\text{det}}$.

Building on these primitives, we formulate the full search space, denoted as $\Omega$, as the topological space of all constructible heterogeneous workflows. It represents the infinite set of all valid directed acyclic graphs derivable from $\mathcal{V}$:
\begin{equation}
    \Omega = \left\{ \mathcal{G} = (\mathcal{V}', \mathcal{E}') \mid \mathcal{V}' \subseteq \mathcal{V}, \mathcal{E}' \subseteq \mathcal{V}' \times \mathcal{V}' \right\},
\end{equation}
where $\mathcal{G} \in \Omega$ represents a structural combination of probabilistic LLM inference and deterministic execution nodes.

\paragraph{Workflow Formulation.}
We formalize the agentic workflow as a parameterized computational graph synthesized by the meta-agent. The synthesis process is defined as:
\begin{equation}
    \mathcal{G}_{\theta} = (\mathcal{V}_{\text{gen}}, \mathcal{E}_{\text{gen}}) \sim \pi_{\theta}(q),
\end{equation}
where $q$ is the input query. The policy $\pi_{\theta}$ synthesizes the set of nodes $\mathcal{V}_{\text{gen}}$ and the data flow edges $\mathcal{E}_{\text{gen}}$.

The execution of an input query $q$ is modeled as a recursive message passing process. For each node $v_k$ in the topological order, its output $o_k$ is computed as:
\begin{equation}
    o_k = \Psi \left( v_k, \{ o_j \mid v_j \in \mathcal{N}_{\text{in}}(v_k) \} \right),
\end{equation}
where $\mathcal{N}_{\text{in}}(v_k)$ denotes the set of immediate predecessors of $v_k$. The execution function $\Psi$ distinguishes the processing logic by invoking model inference if $v_k \in \mathcal{V}_{\text{LLM}}$ or executing the synthesized script if $v_k \in \mathcal{V}_{\text{Code}}$. The final result $y$ is derived from the terminal node's output.

\paragraph{Optimization Objectives.}
The integration of code nodes improves execution reliability and economic efficiency.
To determine the workflow structure that best leverages these benefits, we cast the problem as maximizing an aggregated reward function $R(\mathcal{G})$. We normalize the inference cost and execution latency into efficiency scores to align their scales with performance. For a task distribution $\mathcal{D}$, our objective is:
\begin{equation}
    \begin{gathered}
     \mathcal{G}^* = \underset{\mathcal{G} \in \Omega}{\arg\max} \, R(\mathcal{G}),\\
    s.t.,\ R(\mathcal{G}) = \mathbb{E}_{q \sim \mathcal{D}} \left[ \lambda_1 S_q + \lambda_2 \mathcal{U}_c(C_q) + \lambda_3 \mathcal{U}_t(T_q) \right]
    \end{gathered}
\end{equation}
where $S_q$ represents the performance score for a processed query $q$, $C_q$ is the monetary cost derived from token consumption, and $T_q$ quantifies the end-to-end latency. The functions $\mathcal{U}_c(\cdot)$ and $\mathcal{U}_t(\cdot)$ transform $C_q$ and $T_q$ into normalized utility scores via $\mathcal{U}(x) = (1 + \alpha x)^{-1}$, where $\alpha$ is a scaling factor. The hyperparameters $\lambda_1, \lambda_2, \lambda_3 > 0$ control the preference between performance and efficiency.

\section{Methodology}
As illustrated in Figure~\ref{fig:method}, HyEvo is a self-evolving framework that autonomously constructs heterogeneous agentic workflows. While synthesizing atomic nodes offers superior adaptability, it inevitably results in a larger search space. To navigate this challenge efficiently, HyEvo executes through five sequential actions: (1) initializing the population with a seed workflow; (2) sampling context including a parent and references from the multi-island populations; (3) synthesizing a heterogeneous agentic workflow via a meta-agent using a reflect-then-generate mechanism guided by the evolutionary prompt; (4) evaluating the candidate through a cascaded sandbox protocol; and (5) updating populations to maintain diversity. Specifically, after initialization, the system enters an iterative cycle repeating steps 2 through 5 to continuously refine the solutions until the termination condition is met.

\subsection{Population Initialization}
The process begins by initializing the multi-island populations with a seed workflow. We adopt a simple initialization strategy, using a lightweight workflow with a single LLM node for direct inference. This starting point is designed to fully leverage the exploration capabilities of the evolutionary algorithm, allowing complex heterogeneous patterns to emerge autonomously rather than being dictated by human priors. Formally, the seed workflow is instantiated as:
\begin{equation}
    \mathcal{G}_{\text{init}} = (\{v_{\text{root}}\}, \emptyset),
\end{equation}
where $v_{\text{root}} \in \mathcal{V}_{\text{LLM}}$ is a standard input-output prompting node. Prior to evolution, the seed workflow is executed in the sandbox to obtain the initial reward $R_{\text{init}}$, calculated based on performance, cost, and latency, along with the execution logs $\mathcal{L}_{\text{init}}$. These values are used to initialize the global history set $\mathcal{H}_{\text{global}}$, as well as populate both the local elite archive $\mathcal{A}_k$ and the history set $\mathcal{H}_k$ for every island. Initiating the search from this simplest state allows the multi-objective optimization to drive the autonomous emergence of complexity, including LLM nodes, code nodes, and topologies.

\subsection{Context Sampling}
To effectively guide evolution, we employ a context-aware sampling strategy that provides the meta-agent with a primary parent for improvement and relevant references. At evolutionary iteration $t$, we randomly select an island $k$ and begin the process by selecting a parent workflow $\mathcal{G}_{\text{parent}}$. To balance the trade-off between refining high-performing solutions and exploring new possibilities, we implement a probabilistic selection policy governed by exploration and exploitation ratios, $\rho_{\text{exp}}$ and $\rho_{\text{ploit}}$. The parent selection is formulated as:
\begin{equation}
    \resizebox{0.91\hsize}{!}{$
        \displaystyle
        \mathcal{G}_{\text{parent}} \sim \begin{cases} 
            \text{Uniform}(\mathcal{H}_{k}) & \text{if } r < \rho_{\text{exp}} \\
            \text{Uniform}(\mathcal{A}_{k}) & \text{if } \rho_{\text{exp}} \le r < \rho_{\text{exp}} + \rho_{\text{ploit}} \\
            \text{Uniform}(\mathcal{H}_{\text{global}}) & \text{otherwise}
        \end{cases},
    $}
\end{equation}
where $r \sim \mathcal{U}[0,1]$ is a random factor. Here, $\mathcal{H}_{k}$ denotes the local history set containing all valid programs on island $k$, $\mathcal{A}_{k}$ represents the elite archive of island $k$, and $\mathcal{H}_{\text{global}}$ covers the historical population across all islands. This mechanism ensures that the meta-agent primarily focuses on robust candidates from the archive while maintaining a steady flow of diverse structures through historical and global perturbations.

Once $\mathcal{G}_{\text{parent}}$ is determined, we retrieve its associated error logs $\mathcal{L}_{\text{parent}}$. To form the final context $\mathcal{P}_{\text{ctx}}$, we explicitly construct a reference set containing both the best and diverse instances. Specifically, we rank all workflows in $\mathcal{A}_{k}$ according to their composite rewards $R(\mathcal{G})$, selecting the top-ranked candidate as $\mathcal{G}_{\text{top}}$ to serve as a high-quality exemplar. To prevent mode collapse and encourage variability, we leverage the archive's behavior space to select the diverse reference $\mathcal{G}_{\text{div}}$ from a distinct occupied grid cell. Consequently, the composite evolutionary prompt is formulated as:
\begin{equation}
    \mathcal{P}_{\text{ctx}} = \left\langle (\mathcal{G}_{\text{parent}}, \mathcal{L}_{\text{parent}}), \mathcal{G}_{\text{top}}, \mathcal{G}_{\text{div}} \right\rangle.
\end{equation}

\subsection{Heterogeneous Workflow Synthesis}
The meta-agent $\pi_{\theta}$ synthesizes offspring workflows by leveraging the provided context $\mathcal{P}_{\text{ctx}}$ through a reflect-then-generate mechanism. Unlike random mutation, this approach utilizes LLM reasoning for the joint optimization of topology and node semantics. First, in the reflection phase, the meta-agent performs a comparative analysis to diagnose the shortcomings of the primary parent $\mathcal{G}_{\text{parent}}$ by examining its execution failure logs $\mathcal{L}_{\text{parent}}$. It simultaneously draws insights from the high-performing exemplar $\mathcal{G}_{\text{top}}$ and the diverse reference $\mathcal{G}_{\text{div}}$ to distinguish between effective and ineffective patterns. Based on this comparison, the agent identifies specific structural bottlenecks, such as logical errors or edge-case failures, and formulates a natural language diagnosis $h_{\text{reflect}}$:
\begin{equation}
    h_{\text{reflect}} \sim \pi_{\theta}(\text{analyze} \mid \mathcal{I}_{\text{sys}}, \mathcal{P}_{\text{ctx}}).
\end{equation}

Next, guided by this diagnostic gradient, the generation phase synthesizes a new heterogeneous workflow $\mathcal{G}_{\text{new}}$. The meta-agent defines both the topology and the node parameters to resolve the identified issues:
\begin{equation}
    \mathcal{G}_{\text{new}} \sim \pi_{\theta}(\text{synthesize} \mid \mathcal{I}_{\text{sys}}, \mathcal{P}_{\text{ctx}}, h_{\text{reflect}}).
\end{equation}

Specifically, the meta-agent instantiates the workflow by generating the complete implementation of the topological structure and atomic nodes. For probabilistic components, it drafts context-specific instructions for $v^{\text{LLM}}$ to perform sub-tasks such as problem decomposition or reflective analysis. For deterministic components, it synthesizes executable logic for $v^{\text{Code}}$ to perform precisely defined rule-based tasks, such as format validation or other predefined procedures, ensuring correctness and consistency. Furthermore, the meta-agent specifies the control flow to orchestrate the interaction between these heterogeneous units, utilizing conditional branches to manage data routing effectively.

\subsection{Sandbox Evaluation}

The autonomous synthesis of heterogeneous workflows introduces execution risks such as syntax errors and logical failures. Furthermore, validating every generated candidate on the complete dataset during the evolutionary process would be computationally expensive. To address these challenges, we design a cascaded sandbox evaluation protocol to ensure robust performance estimation while improving the efficiency of the evaluation process. This mechanism employs a hierarchical strategy that rapidly filters out defective candidates with minimal overhead before allocating comprehensive resources to high-potential solutions.

The process begins with a rapid screening on the first half of the validation set, $\mathcal{D}_{\text{part}}$. We execute the workflow in an isolated sandbox environment to capture runtime errors and estimate a preliminary composite reward. Only candidates exceeding a specific reward threshold $\gamma$ proceed to the full evaluation stage. In this phase, the workflow is evaluated on the remaining half of the dataset to precisely quantify objective metrics $M$, including performance score $S_q$, cost $C_q$, and latency $T_q$. We use these results to compute the aggregated reward $R(\mathcal{G})$ for fitness comparison and to collect execution logs $\mathcal{L}$. We also extract the structural features required for the population update. These results are used to identify high-performing candidates and maintain a diverse population.

\subsection{Population Update}
To maintain robust population diversity and prevent premature convergence, we manage the evolved workflows using a multi-island strategy. Each island maintains a local population consisting of two components: a comprehensive historical set $\mathcal{H}_{k}$ that stores all valid workflows, and a structured elite archive $\mathcal{A}_{k}$ that discretizes the phenotypic space into grid cells. Upon completing the evaluation, we map each valid workflow $\mathcal{G}$ to a corresponding grid cell $\mathbf{b}$ based on its behavior descriptors to update $\mathcal{A}_{k}$, while simultaneously appending it to $\mathcal{H}_{k}$. In our implementation, we define the behavior space using workflow complexity, measured by the total node count, and reasoning density, defined as the proportion of LLM nodes. Assuming the reward of an empty cell is $-\infty$, the update rule for the local elite archive $\mathcal{A}_{k}$ is:
\begin{equation}
    \mathcal{A}_{k}(\mathbf{b}) \leftarrow \begin{cases} 
        \mathcal{G} & \text{if } R(\mathcal{G}) > R(\mathcal{A}_{k}(\mathbf{b})) \\
        \mathcal{A}_{k}(\mathbf{b}) & \text{otherwise}
    \end{cases}.
\end{equation}

This mechanism ensures that the archive retains only the highest-performing solution for each specific structural configuration. Furthermore, to prevent the search from getting trapped in local optima, we employ a migration mechanism among $K$ islands. At fixed intervals $\Delta_{\text{mig}}$, each island $k$ copies its top-ranked elites from the local archive $\mathcal{A}_k$ and transfers them to its neighbor. This strategy allows superior structural patterns to spread across the population, helping to maintain diversity and revitalize the evolutionary process.

\subsection{Implementation of HyEvo}

\begin{algorithm}[t]
\caption{HyEvo Framework}
\label{alg:hyevo}
\begin{algorithmic}[1]
\renewcommand{\algorithmicrequire}{\textbf{Input:}}
\renewcommand{\algorithmicensure}{\textbf{Output:}}
\small
\REQUIRE i) $\mathcal{G}_{\text{init}}$, initial seed workflow; ii) $\mathcal{D}$, validation set; iii) $N_{\text{iter}}$, total iterations; iv) $K$, number of islands; v) $\Delta_{\text{mig}}$, migration interval;
\ENSURE $\mathcal{G}_{\text{best}}$, an optimal workflow.

\STATE $(R_{\text{init}}, \mathcal{L}_{\text{init}}) \leftarrow \textsc{CascadedEval}(\mathcal{G}_{\text{init}}, \mathcal{D})$
\STATE $\mathcal{G}_{\text{best}} \leftarrow \mathcal{G}_{\text{init}}$
\STATE $\mathcal{H}_{\text{global}} \leftarrow \{(\mathcal{G}_{\text{init}}, R_{\text{init}}, \mathcal{L}_{\text{init}})\}$

\FOR{$k \leftarrow 1$ \TO $K$}
    \STATE $\mathcal{A}_{k} \leftarrow \{(\mathcal{G}_{\text{init}}, R_{\text{init}}, \mathcal{L}_{\text{init}})\}$
    \STATE $\mathcal{H}_{k} \leftarrow \{(\mathcal{G}_{\text{init}}, R_{\text{init}}, \mathcal{L}_{\text{init}})\}$
\ENDFOR

\FOR{$t \leftarrow 1$ \TO $N_{\text{iter}}$}
    \STATE $k \leftarrow \text{randomSelect}(1, K)$
    \STATE $(\mathcal{G}_{\text{parent}}, \mathcal{L}_{\text{parent}}) \leftarrow \textsc{GetParent}(\mathcal{H}_{k}, \mathcal{A}_{k}, \mathcal{H}_{\text{global}})$
    \STATE $(\mathcal{G}_{\text{top}}, \mathcal{G}_{\text{div}}) \leftarrow \textsc{GetRefs}(\mathcal{A}_{k})$
    \STATE $\mathcal{P}_{\text{ctx}} \leftarrow \langle (\mathcal{G}_{\text{parent}}, \mathcal{L}_{\text{parent}}), \mathcal{G}_{\text{top}}, \mathcal{G}_{\text{div}} \rangle$
    \STATE $\mathcal{G}_{\text{new}} \leftarrow \textsc{ReflectAndGenerate}(\mathcal{P}_{\text{ctx}}, \mathcal{I}_{\text{sys}})$
    \STATE $(R_{\text{new}}, \mathcal{L}_{\text{new}}) \leftarrow \textsc{CascadedEval}(\mathcal{G}_{\text{new}}, \mathcal{D})$
    
    \IF{$R_{\text{new}} \neq \text{NULL}$}
        \STATE \textsc{UpdateArchive}($\mathcal{A}_{k}, \mathcal{G}_{\text{new}}, R_{\text{new}}, \mathcal{L}_{\text{new}}$)
        \STATE $\mathcal{H}_{k} \leftarrow \mathcal{H}_{k} \cup \{(\mathcal{G}_{\text{new}}, R_{\text{new}}, \mathcal{L}_{\text{new}})\}$
        \STATE $\mathcal{H}_{\text{global}} \leftarrow \mathcal{H}_{\text{global}} \cup \{(\mathcal{G}_{\text{new}}, R_{\text{new}}, \mathcal{L}_{\text{new}})\}$
        \IF{$R_{\text{new}} > R(\mathcal{G}_{\text{best}})$}
            \STATE $\mathcal{G}_{\text{best}} \leftarrow \mathcal{G}_{\text{new}}$
        \ENDIF
    \ENDIF
    
    \IF{$t \bmod \Delta_{\text{mig}} == 0$}
        \STATE \textsc{RingMigration}($\{\mathcal{A}_k\}_{k=1}^K$)
    \ENDIF
\ENDFOR
\RETURN $\mathcal{G}_{\text{best}}$
\end{algorithmic}
\end{algorithm}

Algorithm \ref{alg:hyevo} outlines the framework. Lines 1-7 initialize global and local populations. Next, lines 9-12 utilize the \textsc{GetParent} and \textsc{GetRefs} functions to sample contextual information and construct the evolutionary prompt $\mathcal{P}_{\text{ctx}}$ as detailed in Section 4.2. Subsequently, line 13 synthesizes a new heterogeneous workflow via the \textsc{ReflectAndGenerate} described in Section 4.3, which is then assessed by the \textsc{CascadedEval} protocol defined in Section 4.4. If the candidate is valid, the \textsc{UpdateArchive} function in lines 15-22 updates the local elite archive and history sets according to the update rules in Section 4.5. In addition, lines 23-25 execute \textsc{RingMigration} to exchange elite individuals between neighboring islands. Finally, line 27 returns $\mathcal{G}_{\text{best}}$.

\section{Experiment}

In this section, we designed comprehensive experiments to answer the following three Research Questions (RQs):

\paragraph{RQ1 (Superiority)} Does HyEvo outperform state-of-the-art methods on complex reasoning and coding tasks?

\paragraph{RQ2 (Mechanism Necessity)} How do the key components impact the evolutionary process?

\paragraph{RQ3 (Evolutionary Patterns)} How does HyEvo evolve to discover optimal workflows and synergistic hybrid patterns?

\subsection{Experimental Settings}

\paragraph{Datasets.} We evaluated our proposed method on five widely used real-world benchmark datasets, providing a comprehensive assessment across diverse reasoning and coding tasks. The selected datasets cover two primary domains: i) Mathematical Reasoning, including GSM8K~\cite{GSM8K}, MATH~\cite{MATH}, and MultiArith~\cite{MultiArith}, which require complex multi-step logical deduction; and ii) Code Generation, including HumanEval~\cite{humanEval} and MBPP~\cite{MBPP}, which evaluate the system's ability to synthesize functional code solutions. These datasets serve as standard benchmarks for assessing the capability of agentic workflows to handle both structured and unstructured problems.

\paragraph{Baselines.}
We compared HyEvo against two categories of baselines: i) Manually Designed Workflows, ranging from prompting strategies (CoT~\cite{wei2022chain}, ComplexCoT~\cite{fu2022complexity}, SC~\cite{wang2023selfconsistency}) to hand-crafted multi-agent systems (MultiPersona~\cite{multi-persona}, LLM-Debate~\cite{arXiv2023_MultiAgent-Debate}, LLM-Blender~\cite{blender}, DyLAN~\cite{arXiv2023_Dynamic-LLM-Agent}, AgentVerse~\cite{chen2024agentverse}, MacNet~\cite{qian2025scaling}); and ii) Automated Agentic Workflows, including learning-based (GPTSwarm~\cite{zhuge2024gptswarm}, MaAS~\cite{zhangmulti}) and search-based frameworks (AutoAgents~\cite{ijcai2024p3}, ADAS~\cite{huautomated}, AgentSquare~\cite{shangagentsquare}, AFlow~\cite{zhang2025aflow}).

\paragraph{Implementation details.}
We used \texttt{gpt-4o-mini} as the closed-source LLM backbone as well as two open-source models, \texttt{DeepSeek-V3} and \texttt{Qwen3-Max}, all of which are accessed via APIs with the temperature set to $1$. Building on established workflow automation methodologies~\cite{zhang2025aflow,zhangmulti}, we split the data into validation and test sets at a 1:4 ratio. Accordingly, we adopt accuracy as the evaluation metric for mathematical reasoning benchmarks (GSM8K, MATH, MultiArith) and pass@1 for code generation benchmarks (HumanEval, MBPP). For our proposed HyEvo, we set the number of evolutionary islands to $K=2$, the total number of iterations to $N_{\text{iter}}=40$ and the migration interval to $\Delta_{\text{mig}}=15$. For the context sampling strategy, we set the exploration and exploitation ratios as $\rho_{\text{exp}}=0.3$ and $\rho_{\text{ploit}}=0.5$, respectively. Regarding optimization objectives, we prioritized performance by setting the weights to $\lambda_1=0.9$ and $\lambda_2=\lambda_3=0.05$. The scaling factors are set to $\alpha=5$ for monetary cost and $\alpha=1/60$ for latency, effectively normalizing these metrics to comparable utility scores. During each iteration, the system randomly selects one island for candidate generation. We used the performance results reported in \cite{zhangmulti}. For the efficiency analysis, we reproduced AFlow within our experimental environment to ensure a strictly fair comparison of execution latency and token consumption. All experiments were conducted on a Linux server (Ubuntu 20.04) equipped with an Intel Core i9-13900K CPU and 64GB RAM. 

\subsection{Comparison with State-of-the-Arts (RQ1)}

\begin{table}[t]
    \centering
    \scriptsize
    \setlength{\tabcolsep}{3.5pt}
    \begin{tabular}{l|c|c|c|c|c|c}
        \toprule
        \textbf{Method} & \textbf{GSM8K} & \textbf{MATH} & \textbf{M. Arith} & \textbf{H. Eval} & \textbf{MBPP} & \textbf{Avg.} \\
        \midrule
        Vanilla & 87.45 & 46.29 & 96.85 & 87.08 & 71.83 & 77.50 \\
        CoT & 87.10 & 46.40 & 96.31 & 88.13 & 71.83 & 77.95 \\
        ComplexCoT & 86.89 & 46.53 & 96.70 & 87.49 & 72.36 & 78.00 \\
        SC (CoT$\times$5) & 87.57 & 47.91 & 96.58 & 88.60 & 73.60 & 78.85 \\
        \midrule
        MultiPersona & 87.50 & 45.43 & 97.49 & 88.32 & 73.19 & 78.39 \\
        LLM-Debate & 89.47 & 48.54 & 97.33 & 88.68 & 70.29 & 78.86 \\
        LLM-Blender & 88.35 & 45.92 & 97.29 & 88.80 & 77.05 & 79.68 \\
        DyLAN & 89.98 & 48.63 & 97.12 & 90.42 & 77.30 & 80.69 \\
        AgentVerse & 89.91 & 47.35 & 97.50 & 89.29 & 74.28 & 79.67 \\
        MacNet & 87.95 & 45.18 & 96.03 & 84.57 & 65.28 & 75.00 \\
        \midrule
        GPTSwarm & 89.14 & 47.88 & 96.79 & 89.32 & 77.43 & 80.11 \\
        MaAS & 92.30 & 51.82 & 98.80 & 92.85 & 82.17 & 83.59 \\
        \midrule
        AutoAgents & 87.69 & 45.32 & 96.42 & 87.64 & 71.95 & 77.80 \\
        ADAS & 86.12 & 43.18 & 96.02 & 84.19 & 68.13 & 75.13 \\
        AgentSquare & 87.62 & 48.51 & 97.77 & 89.08 & 78.46 & 80.29 \\
        AFlow & 91.16 & 51.28 & 96.22 & 90.93 & 81.67 & 82.25 \\
        \textbf{HyEvo (Ours)} & \textbf{93.36} & \textbf{53.91} & \textbf{99.67} & \textbf{93.89} & \textbf{83.28} & \textbf{84.82} \\
        \bottomrule
    \end{tabular}
    \caption{Performance (\%) comparison with state-of-the-art baselines on five benchmarks, where \texttt{M.Arith} and \texttt{H.Eval} stand for MultiArith and HumanEval, respectively.}
    \label{tab:performance_comparison}
\end{table}

To demonstrate the superiority of HyEvo, we systematically compared our approach with three categories of state-of-the-art baselines across both reasoning and coding benchmarks, using \texttt{gpt-4o-mini} as a unified backbone for fair comparison. Table~\ref{tab:performance_comparison} presents the comparison results. Columns 2-6 report the experimental results on specific datasets, while the last column shows the average performance. This table shows that HyEvo consistently achieves the best performance across all benchmarks. Overall, HyEvo achieves an average score of 84.82\%, surpassing the leading baselines MaAS and AFlow by 1.23\% and 2.57\%, respectively. These results fully demonstrate the versatility and superiority of HyEvo.

\begin{table}[t]
    \centering
    \small
    \scriptsize
    \setlength{\tabcolsep}{3.5pt}
    \begin{tabular}{l|c|cc|cc|cc}
        \toprule
        \multirow{3}{*}{\textbf{Dataset}} & \multirow{3}{*}{\textbf{Metric}} & \multicolumn{2}{c|}{gpt-4o-mini} & \multicolumn{2}{c|}{DeepSeek-V3} & \multicolumn{2}{c}{Qwen3-Max} \\
        \cmidrule(lr){3-4} \cmidrule(lr){5-6} \cmidrule(lr){7-8}
         & & AFlow & HyEvo & AFlow & HyEvo & AFlow & HyEvo \\
        \midrule
        \multirow{3}{*}{MATH} 
         & Perf. $\uparrow$ & 51.65 & \textbf{53.91} & 67.49 & \textbf{71.60} & 82.30 & \textbf{84.98} \\
         & Cost $\downarrow$ & 3.78 & \textbf{1.85} & 2.75 & \textbf{0.93} & 2.04 & \textbf{0.50} \\
         & Time $\downarrow$ & 138.86 & \textbf{30.76} & 203.75 & \textbf{92.57} & 236.08 & \textbf{83.35} \\
        \midrule
        \multirow{3}{*}{MBPP} 
         & Perf. $\uparrow$ & 81.52 & \textbf{83.28} & 83.28 & \textbf{86.51} & 83.87 & \textbf{85.34} \\
         & Cost $\downarrow$ & 1.05 & \textbf{0.08} & 0.49 & \textbf{0.03} & 0.38 & \textbf{0.02} \\
         & Time $\downarrow$ & 23.93 & \textbf{2.42} & 38.03 & \textbf{2.36} & 57.14 & \textbf{3.85} \\
        \bottomrule
    \end{tabular}
    \caption{Efficiency analysis on complex reasoning and coding tasks. We report Performance, Average Cost ($10^{-3}$ USD), and Average Latency (s). \texttt{Perf.} stands for Performance.}
    \label{tab:combine_comparison}
\end{table}

In addition to achieving high performance, the workflows evolved by HyEvo demonstrate superior resource efficiency, significantly reducing inference costs and execution latency across different LLM backbones. Table~\ref{tab:combine_comparison} presents the efficiency analysis on the computationally demanding MATH and MBPP datasets. As shown in the top headers, the comparison is conducted across three distinct LLM backbones. The second column lists the evaluation metrics, including Performance, Cost ($10^{-3}$ USD), and Latency (s), while columns 3-8 report the corresponding results for AFlow and HyEvo. We selected AFlow as the primary baseline for comparison because it is a state-of-the-art method with fully reproducible implementations. We specifically targeted MATH and MBPP for this efficiency stress test because they are computationally demanding and require extensive multi-step reasoning.

From Table~\ref{tab:combine_comparison}, we can find that HyEvo improves both performance and resource efficiency. In terms of monetary cost, HyEvo consistently demonstrates significant savings across all configurations. Specifically, it reduces expenses by approximately 2 to 4 times on the MATH dataset and achieves a substantial reduction of 13 to 19 times on the MBPP dataset. Regarding execution latency, HyEvo accelerates inference by 2 to 5 times on MATH and 10 to 16 times on MBPP. This efficiency gain is primarily attributed to our hybrid design, which replaces redundant LLM reasoning with atomic code nodes, effectively eliminating the latency and token consumption associated with repetitive inference.

\subsection{Mechanism Necessity (RQ2)}

\begin{figure}[t]
    \centering
    \includegraphics[width=0.8\linewidth]{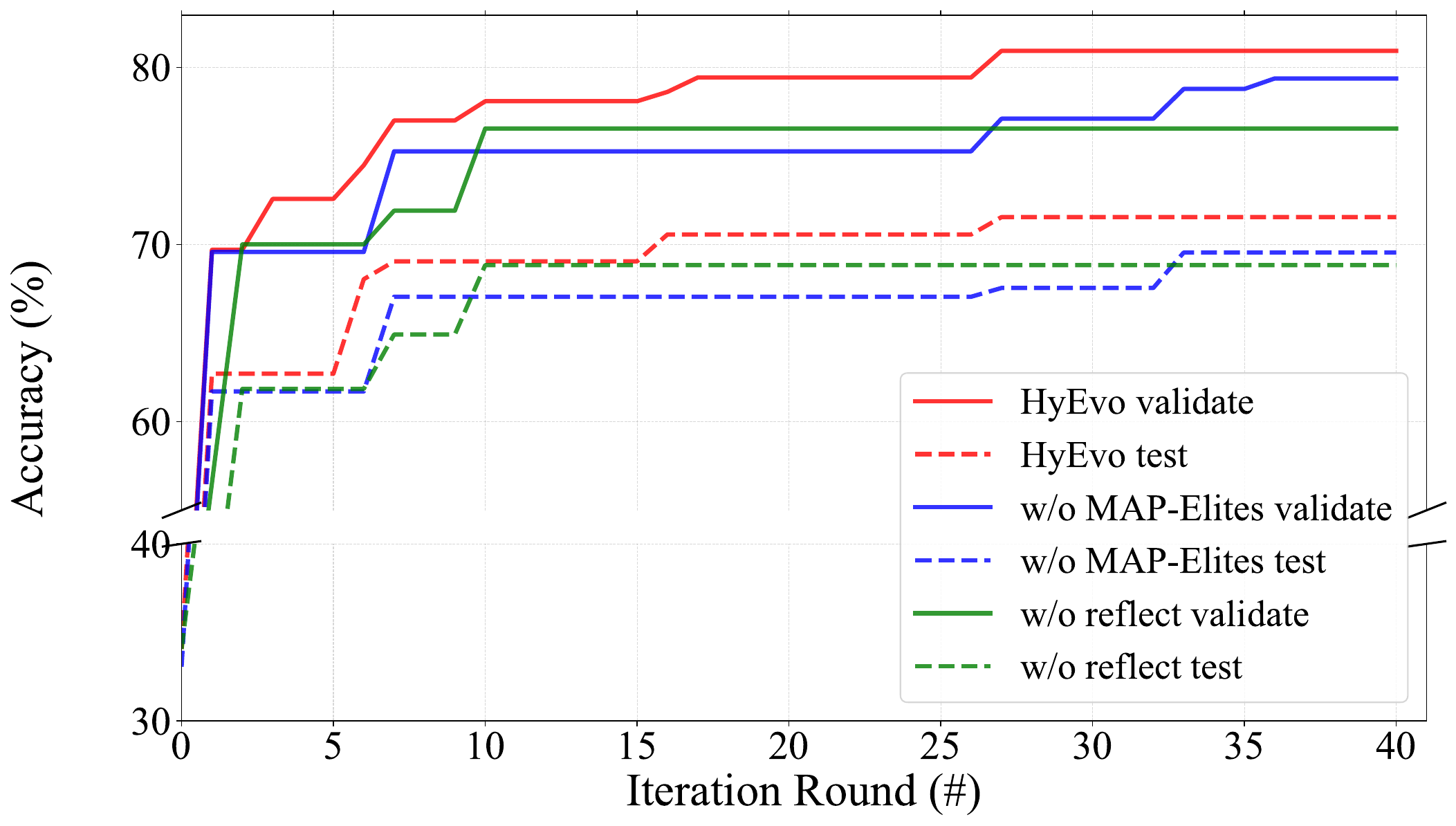}
        \vspace{-0.1in}
    \caption{Ablation study of HyEvo components.}
    \vspace{-0.1in}
    \label{fig:comparison}
\end{figure}

To validate the necessity of key components, we conducted ablation studies on the MATH dataset using \texttt{DeepSeek-V3} as the backbone, focusing on the reflect-then-generate mechanism and the MAP-Elites strategy. The validation and test accuracy curves over 40 iterations are presented in Figure~\ref{fig:comparison}.

\begin{figure*}[t]
    \centering
    \includegraphics[width=0.95\linewidth]{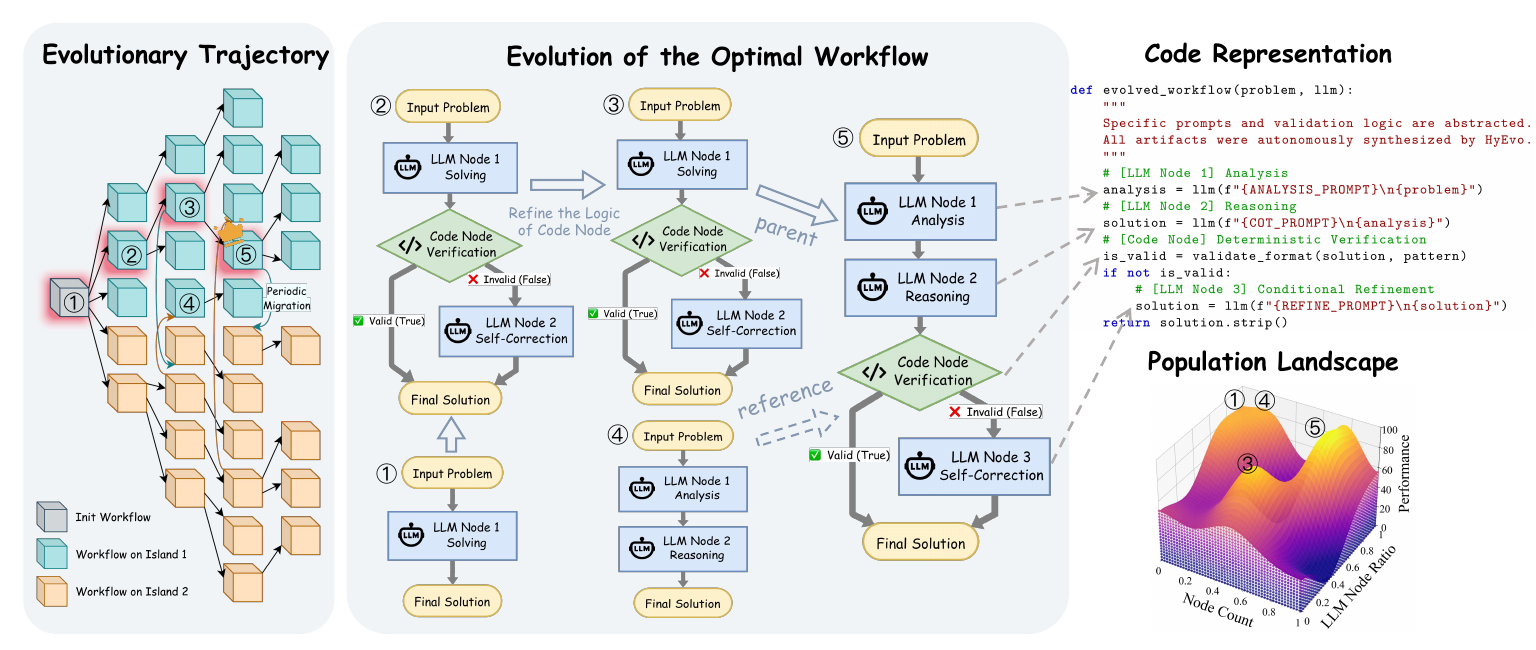}
    \vspace{-0.1in}
    \caption{The evolutionary trajectory of the optimal hybrid workflow on the \texttt{MATH} dataset.}
    \vspace{-0.1in}
    \label{fig:case_study}
\end{figure*}

\paragraph{Impact of Reflection Mechanism.} 
As shown in Figure~\ref{fig:comparison}, the variant without reflection (\textit{w/o} Reflect) exhibits premature convergence. Its validation accuracy increases rapidly in the early stages, reaching 76.47\% within the first 10 iterations, after which no further improvement occurs. In contrast, the full HyEvo framework continues to make steady progress beyond this point, achieving a final validation accuracy of 80.67\%, yielding an absolute improvement of 4.20 percentage points over the no-reflection variant. This result demonstrates that the reflection mechanism effectively helps the evolutionary process escape local optima by correcting suboptimal workflow structures. The same improvement trend is observed on the test set, indicating that reflection enhances generalization rather than overfitting.

\paragraph{Impact of MAP-Elites Strategy.}
Compared to the greedy selection baseline (\textit{w/o} MAP-Elites), the quality-diversity search in HyEvo demonstrates both faster convergence and higher final accuracy. As illustrated in Figure~\ref{fig:comparison}, the greedy variant plateaus at 78.99\% validation accuracy, whereas HyEvo continues to improve and reaches 80.67\% at convergence. The resulting performance gap of 1.68 percentage points highlights the importance of maintaining population diversity during evolutionary search. Without an elite archive to preserve structurally distinct candidates and provide diverse references for the reflection mechanism, the search quickly collapses to a narrow region of the space, limiting further exploration and leading to suboptimal solutions. This advantage is consistently observed on the test set, confirming that the MAP-Elites strategy improves both optimization stability and final generalization performance.

\subsection{Evolutionary Patterns (RQ3)}

To understand why HyEvo discovers superior workflows, Figure~\ref{fig:case_study} presents a detailed case study of the evolutionary process, which visualizes the optimization trajectory from four interconnected perspectives. The evolutionary trajectory (left) traces the solution path across the islands, highlighting the ancestry of optimal workflow. Corresponding to these highlighted milestones, the evolution of the optimal workflow (middle) details their specific functional and topological progression. The code representation (top right) shows simplified executable code for the final optimal workflow. Finally, the population landscape (bottom right) illustrates the fitness distribution within the elite archive, providing statistical evidence for the superiority of hybrid structures.

As illustrated in the evolutionary trajectory, the process begins with Workflow 1, a simple linear chain in which a single LLM directly generates the answer. Driven by the multi-island evolutionary strategy, the isolated populations naturally drifted toward distinct optimization directions. On Island 1, the search process favored strict output control. This preference led to the evolution of Workflow 2, which features deterministic code verification, and then to its improvement into robust Workflow 3. Meanwhile, the population on Island 2 focused on deeper reasoning structures. This trend led to Workflow 4, which decomposes the task into sequential analysis and solution steps. Crucially, the periodic migration mechanism subsequently bridged these distinct paths by transferring Workflow 4 from Island 2 to Island 1. Leveraging this transfer, the meta-agent on Island 1 synthesized the strengths of both paths via the reflect-then-generate mechanism. It selected locally optimized Workflow 3 as the parent and migrated Workflow 4 as the reference to generate the superior Workflow 5. Consequently, this final workflow integrates these complementary traits by inheriting the deep reasoning topology from the migrated workflow and the rigorous format verification logic from the local path.

The final evolved workflow demonstrates a highly efficient division of labor by assigning flexible reasoning to LLM nodes and strict format verification to a code node. Additionally, it integrates a conditional self-correction process that triggers refinement only when format constraints are violated, ensuring high accuracy without wasting tokens on simple queries. This specific synergistic advantage is supported by the population landscape shown in the bottom-right. By mapping the fitness distribution, the landscape reveals that performance does not scale linearly with node count; instead, high-performing solutions emerge within the hybrid regions. This distribution confirms that relying solely on pure LLM or rigid code methods is insufficient, validating that our hybrid evolutionary strategy is essential for discovering efficient solutions with superior performance.

\section{Conclusion}

In this paper, we propose HyEvo, a self-evolving framework that employs a heterogeneous atomic synthesis paradigm. Unlike existing methods confined to homogeneous workflows, where all task-level computation relies on probabilistic LLM inference through predefined operators, HyEvo autonomously constructs synergistic architectures by integrating atomic LLM nodes for semantic reasoning and code nodes for deterministic execution. To efficiently navigate the expansive hybrid search space, we introduce an LLM-driven multi-island evolutionary strategy empowered by a reflect-then-generate mechanism. Comprehensive experiments demonstrate that HyEvo outperforms state-of-the-art baselines, achieving exceptional performance while reducing inference costs and latency by up to an order of magnitude.

\clearpage
\bibliographystyle{named}
\bibliography{ijcai26}

\end{document}